%% file: main.tex
\documentclass[10pt,twocolumn,letterpaper]{article}

\usepackage[pagenumbers]{cvpr}

\usepackage[T1]{fontenc}
\usepackage{amsthm}
\usepackage{tikz}
\usetikzlibrary{arrows.meta}
\usepackage{pgfplots}
\pgfplotsset{compat=1.16}

\definecolor{linkblue}{rgb}{0.21,0.49,0.74}
\usepackage[pagebackref,breaklinks,colorlinks,allcolors=linkblue]{hyperref}
\hypersetup{pdftitle={Exact Sparse Assignment with Costly Target Embeddings: Fixed-Realization Certificates and an Empirical Audit},
  pdfauthor={Hamidreza Hasani Balyani, Mahtab Movahhedrad, Seyed Pouyan Mousavi Davoudi}}

\newtheorem{theorem}{Theorem}
\newtheorem{proposition}{Proposition}
\newtheorem{remark}{Remark}
\input{generated/macros.tex}

\makeatletter
\newcommand{\blfootnote}[1]{\begingroup\renewcommand\thefootnote{}\footnote{#1}\addtocounter{footnote}{-1}\endgroup}
\makeatother

\title{Exact Sparse Assignment with Costly Target Embeddings:\\
Fixed-Realization Certificates and an Empirical Audit}

\author{Hamidreza Hasani Balyani\textsuperscript{1}\qquad
Mahtab Movahhedrad\textsuperscript{2}\qquad
Seyed Pouyan Mousavi Davoudi\textsuperscript{3}\\[0.4em]
{\normalsize\textsuperscript{1}Independent Researcher\qquad
\textsuperscript{2}University of Southern California\qquad
\textsuperscript{3}Islamic Azad University, Amol Branch}}

\begin{document}
\maketitle

\begin{abstract}
\blfootnote{This work has been submitted to the IEEE for possible publication. Copyright may be transferred without notice, after which this version may no longer be accessible.}%
Embedding-based one-to-one association appears throughout vision and data integration: detections are
associated with tracks, items are matched across catalogues, and records are linked across databases.
Such pipelines often solve a sparse optional assignment whose edge scores add a cheap term to an inner
product of neural embeddings. When target embeddings are costly, one can acquire them selectively and stop
once a certificate proves that the current assignment is already optimal. We give such a certificate for
unit-norm embeddings. It bounds the unknown part of every reassignment by a support function and
maximizes the resulting robust gains with a second sparse assignment, and a shared owner-centred bound
dominates any fixed-centre triangle bound. We make precise what is certified when network features are
not batch-invariant: the realized values of the acquired rows and every unit-ball completion of the rest,
not a second forward pass in a different batch. Auditing against a separately batched cache tests a
different property, and on our first pilot that mistake produced a spurious $2.49\times10^{-6}$ objective
failure. We audit the corrected protocol on four public text entity-matching benchmarks with two frozen
encoders; no image data is used. All 96 certificate simulations and all 120 online selective timings were
exact on their own realization (largest objective gap $6.82\times10^{-13}$). At group size 8 the shared
certificate acquires on average 60\% of candidate-active targets, against 74\% (mean centre), 69\% (MEB)
and 60\% (zero centre): it beats the mean and MEB centres but not the zero centre, and savings
concentrate in one benchmark (7\% acquired). The prespecified runtime
claim is contradicted: selective execution is slower by +0.510\,s per run (mean over four datasets;
dataset-cluster 95\% interval [+0.248, +0.772]).
Uniform-microbatch re-encodings preserved the optimum in only 5 of 8 encoder--dataset cells, and 45 of 120
online runs disagreed with an eager cache at near-ties.
\end{abstract}

\section{Introduction}\label{sec:intro}
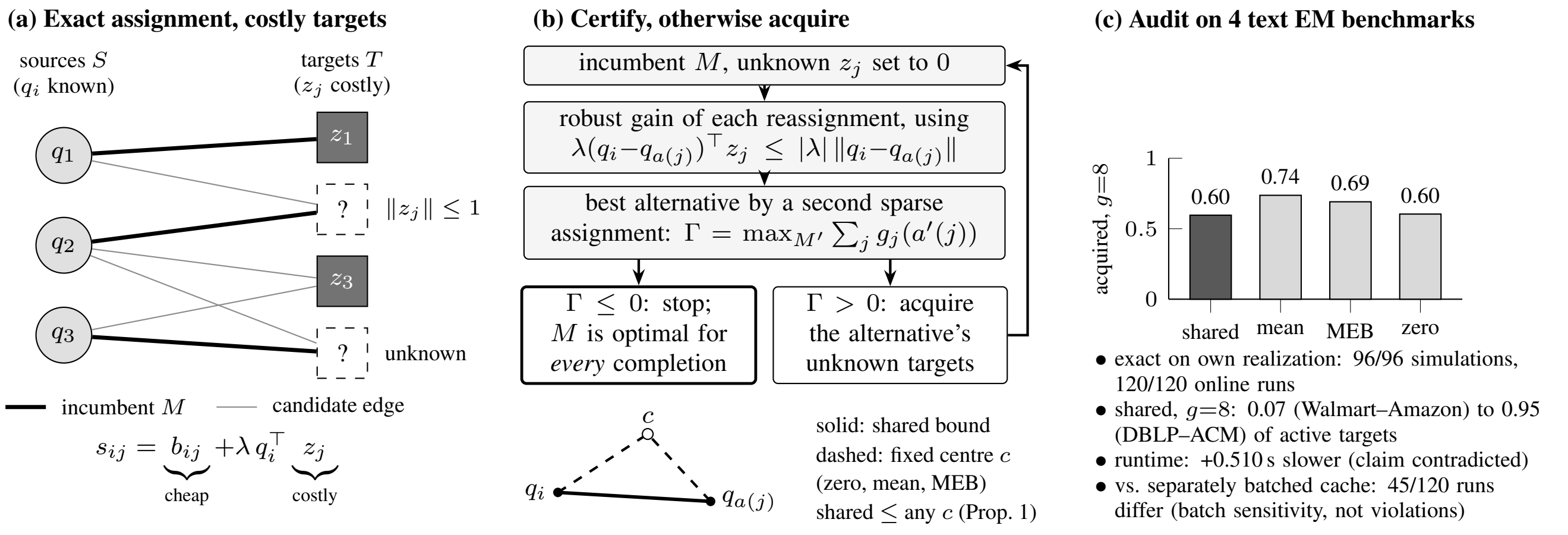
\begin{figure*}[t]\centering
\input{figures/fig1.tex}
\caption{Overview. (a)~A sparse optional assignment whose edge scores add a cheap term $b_{ij}$ to an
inner product with a costly target embedding $z_j$; only some targets have been acquired. (b)~The
certificate bounds the unknown part of every reassignment over the unit ball and maximizes the resulting
robust gains with a second sparse assignment; if the maximum $\Gamma$ is at most zero the incumbent is
optimal for every completion, otherwise the targets in the maximizing alternative are acquired. Bottom:
the shared owner-centred bound (solid) is never looser than the triangle bound through a fixed centre
$c$ (dashed) used by global-centre baselines. (c)~Audit on four public text entity-matching benchmarks:
mean fraction of candidate-active targets acquired at group size 8 (\cref{tab:acq}), exactness on the
certified realization, dataset dependence of the savings, the negative runtime result, and the
cross-batch diagnostic (\cref{sec:numerics}). No image data is used.}\label{fig:teaser}
\end{figure*}

Many matching pipelines score candidate pairs with a learned embedding model and then select a
one-to-one matching. In vision, tracking-by-detection associates detections with existing tracks by an
assignment over cheap geometric cues and appearance-embedding affinities~\citep{bewley2016sort,wojke2017deepsort};
there, track updates usually need every detection's embedding anyway, so whether lazy acquisition helps is
open. In data integration, entity resolution links records across databases in the same
way~\citep{mudgal2018deep,li2020ditto}. When the encoder is expensive relative to the
assignment solve, a natural idea is to acquire target embeddings lazily. Start from the cheap scores,
compute a provisional assignment, and request embeddings only for targets whose unknown features could
still change the optimum (\cref{fig:teaser}). This combines exact assignment~\citep{kuhn1955hungarian}
with a robust-optimization treatment of the unknown features~\citep{bentalnemirovski2002robust}, and is
related in spirit to threshold-style early stopping in top-$k$ retrieval~\citep{fagin2003optimal} and to
active feature acquisition~\citep{saartsechansky2009active}.

The mathematics below applies to any assignment whose edge scores have the form of \cref{eq:score}, with
bounded-norm costly embeddings, whatever the modality. The empirical audit, however, uses only four text
entity-matching benchmarks and two frozen text encoders. We ran no vision experiment and make no claim
about image matching or tracking.

This paper makes four contributions.
\begin{enumerate}
\item A fixed-realization certificate for sparse optional assignment with unit-ball target
embeddings, computable with one additional sparse assignment (\cref{thm:cert}).
\item A shared owner-centred support bound that is never looser than the bound from a global centre:
zero, mean, or an approximate minimum enclosing ball~\citep{badoiu2003smaller}
(\cref{prop:dominance}).
\item A precise statement of the numerical object being certified when transformer features vary
with batch composition, together with a same-realization audit that tests the theorem and a separate
cross-batch audit that tests numerical portability (\cref{sec:numerics}).
\item A four-dataset, two-encoder evaluation that reports exactness, acquisition, runtime
and batch sensitivity, including negative findings (\cref{sec:results}).
\end{enumerate}
The results are mixed, and we report them as such. Exactness holds without exception on the certified
realization. Feature savings are real but strongly dataset-dependent: large on one benchmark, negligible on
two. The prespecified runtime comparison is negative, partly because its geometric schedule acquires almost
every target. On these benchmarks the neural score adds no held-out F1 on average over a tuned lexical
score. We report feature counts and latency separately and never present the former as the latter.

\section{Problem}\label{sec:problem}
Let $G=(S\cup T,E)$ be a frozen sparse bipartite candidate graph produced by a blocker. Each source
$i\in S$ and target $j\in T$ may be used at most once, and unmatched records contribute zero. An edge
$(i,j)\in E$ has score
\begin{equation}\label{eq:score}
s_{ij} = b_{ij} + \lambda\, q_i^\top z_j ,
\end{equation}
where $b_{ij}$ collects the cheap lexical score, a selected threshold offset and a deterministic
$10^{-10}$ tie term. $q_i$ is a known source embedding, and $z_j$ is a target embedding that is costly to
acquire. We assume $\|z_j\|_2\le 1$. An optional assignment is a matching $M\subseteq E$ with value
$V(M)=\sum_{(i,j)\in M}s_{ij}$. We solve it exactly by adding a private zero-score dummy column per
source and calling a sparse shortest-augmenting-path assignment solver~\citep{jonker1987shortest}.

At any time a set $K\subseteq T$ of targets is known. Writing $a(j)$ for the owner of $j$ in an incumbent
$M$ (with $a(j)=0$ and $q_0:=0$ when $j$ is unmatched), the question is whether $M$ is optimal for
\emph{every} completion $\{z_j:\ j\notin K,\ \|z_j\|\le1\}$.

\section{Certificate}\label{sec:cert}
For an unknown target $j$ and a candidate owner $i$ (including the dummy), reassigning $j$ changes the
unknown contribution by $\lambda(q_i-q_{a(j)})^\top z_j\le |\lambda|\,\|q_i-q_{a(j)}\|_2$ by the
Cauchy--Schwarz inequality, with equality attainable by a unit vector. Define the robust gain
$g_j(a(j))=0$ and, for $i\neq a(j)$,
\begin{equation}\label{eq:gain}
g_j(i)=
\begin{cases}
s_{ij}-s_{a(j)j}, & j\in K,\\
b_{ij}-b_{a(j)j}+|\lambda|\,\|q_i-q_{a(j)}\|_2, & j\notin K,
\end{cases}
\end{equation}
for edges $(i,j)\in E$ and for $i=0$ (dropping $j$), with $s_{0j}=b_{0j}=0$.

\begin{theorem}[Fixed-realization certificate]\label{thm:cert}
Let $\Gamma=\max_{M'}\sum_{j} g_j(a'(j))$, where $a'$ ranges over the owner maps of optional
assignments $M'$ in $G$. If $\Gamma\le 0$, then $V(M)\ge V(M')$ for every optional assignment $M'$ and
every unit-ball completion of the unknown targets.
\end{theorem}
\begin{proof}
For any $M'$ with owner map $a'$, $V(M')-V(M)=\sum_{j\in T}\big(s_{a'(j)j}-s_{a(j)j}\big)$ because both
values decompose over targets and unmatched targets contribute zero. For $j\in K$ the term equals
$g_j(a'(j))$. For $j\notin K$ it equals
$b_{a'(j)j}-b_{a(j)j}+\lambda(q_{a'(j)}-q_{a(j)})^\top z_j\le g_j(a'(j))$ for every $\|z_j\|\le1$. Hence
$V(M')-V(M)\le\sum_j g_j(a'(j))\le\Gamma\le 0$. The maximization over $a'$ is itself an optional
assignment on the gain matrix, since $a'$ must use each source at most once.
\end{proof}
The guarantee conditions on the realized source rows and the realized values of the acquired target
rows. It is a statement about one numerical instance.

\paragraph{Global-centre baselines.}
For any fixed centre $c$, $\|q_i-q_{a(j)}\|\le\|q_i-c\|+\|q_{a(j)}-c\|$. Replacing the shared term by this
triangle bound gives a valid but looser certificate. We use $c=0$, the arithmetic mean of owner
vectors, and a deterministic B\u{a}doiu--Clarkson approximate minimum-enclosing-ball
(MEB) centre~\citep{badoiu2003smaller}.

\begin{proposition}[Dominance at a fixed known set]\label{prop:dominance}
For the same incumbent and known set, every robust gain under the shared bound is at most the
corresponding gain under any fixed-centre bound. Hence $\Gamma_{\mathrm{shared}}\le\Gamma_c$, and
whenever the centre-$c$ certificate certifies, the shared certificate does as well.
\end{proposition}
A short proof is given in \cref{sec:s-proof}.
The proposition compares bounds at a fixed state. It does not imply that the shared method acquires
fewer targets along an adaptive trajectory, because different bounds request different targets. That
comparison is empirical.

\begin{remark}[No worst-case savings]
Savings cannot be guaranteed in the worst case. If two sources compete for one target and their lexical
scores tie, then $g_j\ge|\lambda|\,\|q_1-q_2\|>0$ until $z_j$ is known, so no certificate can stop
without acquiring $j$. Replicating this gadget across all targets forces full acquisition.
\end{remark}

\paragraph{Acquisition rule.}
The nominal incumbent sets every unknown $z_j=0$. When $\Gamma>0$, the maximizing alternative (the
\emph{witness}) identifies unknown targets whose owners change. Those targets are requested first,
ordered by their robust-gain priority, and the group is filled to the requested size by priority. The
rule stops as soon as the certificate holds or all candidate-active targets are known, in which case the
exact full scores are used. Group sizes are fixed or geometric (start at $g$ and double per round).

\section{The numerical object being certified}\label{sec:numerics}
A frozen transformer is a deterministic function only up to floating-point execution order. Under FP16
inference, encoding the same text in different batch compositions changes coordinates by up to
$\approx10^{-3}$ (\cref{tab:batch}). We therefore define the feature oracle explicitly. It consists
of the frozen revision, tokenizer and text template, FP16 inference, FP64 direction normalization, FP32
storage with a conservative downward norm correction (so that every row lies in the closed unit ball
checked in FP64), and FP64 score and objective accumulation with outward rounding of robust terms.

\paragraph{Same-realization audit (correctness gate).}
After a timed selective run, the acquired rows are kept bit-for-bit. Only the missing candidate-active
targets are encoded, and the full optional assignment is solved on this completed array. The run is
exact if the reference objective exceeds the selective objective by at most
$10^{-7}+10^{-12}\max(1,|V_{\mathrm{ref}}|,|V_{\mathrm{sel}}|)$. This is exactly the instance quantified
by \cref{thm:cert}, since the completion is one admissible unit-ball completion.

\paragraph{Cross-batch audit (diagnostic).}
Our first pilot instead compared the incumbent against an eagerly encoded full cache produced in a
different batch composition. That cache changes the acquired rows themselves, so a gap there measures
whether the assignment is batch-invariant, not whether the certificate is valid. On that pilot the
zero-centred method showed a gap of $2.4886\times10^{-6}$ in all five repetitions (allowed $10^{-7}$),
and the run was stopped by protocol. A corrected pilot reproduces the cross-batch gap
(\PilotCacheGap) on the same variants while every same-realization audit is exact. This identifies the
original failure as a batch-sensitivity effect on a near-tie, not a certificate violation. We report the
cross-batch comparison and re-encodings at fixed microbatches as a separate numerical-portability result.

\section{Experimental protocol}\label{sec:protocol}
\paragraph{Data and encoders.}
We use the Abt--Buy, Amazon--Google, Walmart--Amazon and DBLP--ACM text entity-matching
benchmarks~\citep{kopcke2010evaluation,mudgal2018deep} from pinned, byte-hashed snapshots. The encoders
are BGE-small-en-v1.5~\citep{xiao2024cpack} and all-MiniLM-L6-v2~\citep{wang2020minilm,reimers2019sentencebert}
at fixed model revisions, with a maximum length of 256 tokens. Each dataset is split 20/80 into development and
test sets by gold connected components, so no entity crosses the split.

\paragraph{Blocking and tuning.}
A character 3--5-gram TF-IDF blocker is used, taking the smallest prespecified width that reaches 0.99
development candidate recall. The embedding weight $\lambda\in\{0.25,0.5,1,1.5\}$ and threshold are
chosen on development data only, and a lexical-only threshold is selected separately. Test labels choose
nothing.

\paragraph{Methods.}
Eager full encoding is the reference. The selective methods are the shared, mean-centred, MEB-centred
and zero-centred certificates. Simulations use one frozen embedding matrix with groups of 8, 32 and 128
and witness ordering. Online timings run actual batched FP16 inference with five repetitions in a
balanced cyclic order. They cover two scopes: \emph{known-source} (source embeddings given, as in the
formal problem) and \emph{end-to-end} (source encoding charged). The prespecified primary runtime
contrast is shared, end-to-end, geometric from 128 versus full, end-to-end.
\emph{Disclosure:} this primary variant was declared after the first pilot showed a slow fixed-32
schedule, so it is not blind to that single observation.

\paragraph{Statistics.}
Contrasts are paired and bootstrapped over dataset-level clusters (10{,}000 draws). Models, groups and
repetitions nested within a dataset are not treated as independent. With four datasets these intervals
are coarse, and we read them descriptively. Hash prefixes of the protocol, configurations and run
records are given in \cref{sec:s-repro}.

\section{Results}\label{sec:results}
\subsection{Exactness}
Every one of the 96 simulations (96 exact) and 120 online selective timings (120 exact) was audited on
its own realization; the largest same-realization objective gap was $6.82\times10^{-13}$, at
floating-point round-off. Against a separately batched eager cache, 45 of 120 online timings would have
\emph{failed} a cache-based audit (largest gap $2.49\times10^{-6}$; \cref{tab:exact}). These are the
cross-batch effects discussed in \cref{sec:numerics}; they are not certificate violations.

\begin{table}[t]\centering\footnotesize
\caption{Online selective timings: same-realization exactness (the gate) versus a separately batched
cache comparison (diagnostic), pooled over datasets and encoders. Scope: e2e = end-to-end,
known = known-source; all three variants use the geometric schedule.}\label{tab:exact}
\begin{tabular}{llrrrr}\toprule
 & & & & exact & cache\\
method & scope & group & runs & (same) & mismatch\\\midrule
shared & e2e   & 128 & 40 & 40 & 15\\
shared & known & 128 & 40 & 40 & 15\\
zero   & known & 128 & 40 & 40 & 15\\
\bottomrule\end{tabular}\end{table}

\begin{table*}[t]\centering\small
\caption{Held-out quality of the eager full assignment (the object every exact selective method
reproduces). $\lambda$ and threshold chosen on development data; lexical-only threshold selected
separately. $\Delta$F1 is computed before rounding.}\label{tab:quality}
\begin{tabular}{llrrrrrr}\toprule
dataset & encoder & cand.\ recall & $\lambda$ & thr. & F1 full & F1 lexical & $\Delta$F1\\\midrule
Abt--Buy & BGE-small & 1.000 & 0.25 & 0.35 & 0.942 & 0.939 & +0.003\\
Abt--Buy & MiniLM-L6 & 1.000 & 0.25 & 0.35 & 0.944 & 0.939 & +0.004\\
Amazon--Google & BGE-small & 0.999 & 0.5 & 0.8 & 0.694 & 0.683 & +0.010\\
Amazon--Google & MiniLM-L6 & 0.999 & 0.5 & 0.8 & 0.696 & 0.683 & +0.013\\
Walmart--Amazon & BGE-small & 1.000 & 0.25 & 0.8 & 0.676 & 0.693 & -0.017\\
Walmart--Amazon & MiniLM-L6 & 1.000 & 0.25 & 0.8 & 0.682 & 0.693 & -0.011\\
DBLP--ACM & BGE-small & 1.000 & 0.5 & 0.65 & 0.985 & 0.987 & -0.002\\
DBLP--ACM & MiniLM-L6 & 1.000 & 0.5 & 0.65 & 0.990 & 0.987 & +0.003\\
\bottomrule\end{tabular}\end{table*}

\subsection{Task quality and blocking}
Averaged within dataset, the neural score changes held-out F1 by +0.000 relative to the separately tuned
lexical-only assignment (dataset-cluster bootstrap 95\% interval [-0.010, +0.009], four clusters;
\cref{tab:quality}). Minimum test candidate recall is 0.999; quality conclusions are restricted to the
frozen candidate graph.

\subsection{Acquisition}
\begin{figure*}[t]\centering
\includegraphics[width=\linewidth]{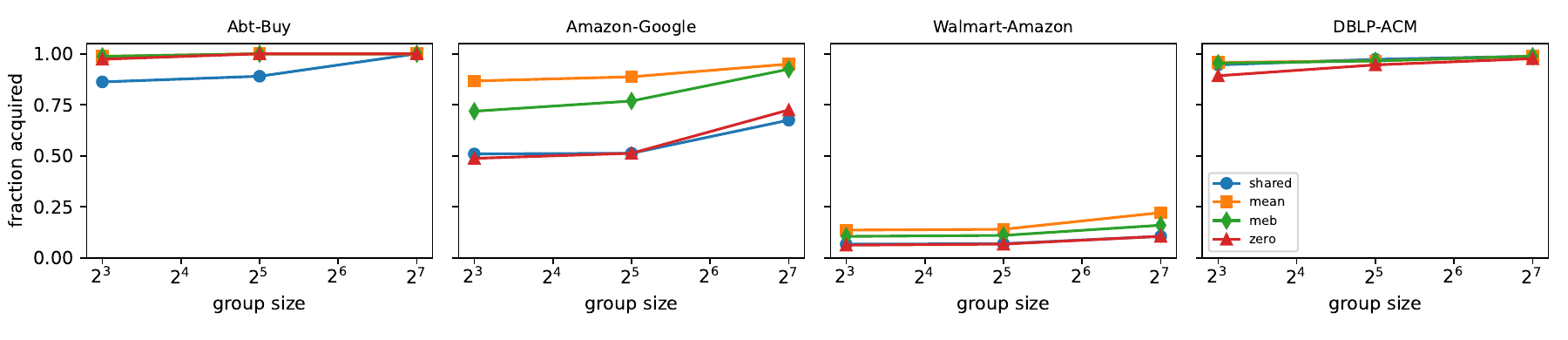}
\caption{Fraction of candidate-active targets acquired before certification by group size
(simulations, witness order, mean over two encoders).}\label{fig:acq}
\end{figure*}

\Cref{tab:acq,fig:acq} report the fraction of candidate-active target embeddings acquired before
certification. Per dataset, the shared certificate acquires Abt--Buy 0.86 at 8, 0.89 at 32, 1.00 at 128;
Amazon--Google 0.51 at 8, 0.51 at 32, 0.67 at 128; Walmart--Amazon 0.07 at 8, 0.07 at 32, 0.11 at 128;
DBLP--ACM 0.95 at 8, 0.97 at 32, 0.99 at 128 of the active targets.

\begin{table}[t]\centering\footnotesize
\caption{Mean fraction of candidate-active target embeddings acquired before certification
(simulations, witness order), averaged over the eight dataset--encoder cells, and the
shared-minus-best-global contrast with a dataset-cluster bootstrap 95\% interval.}\label{tab:acq}
\setlength{\tabcolsep}{4.5pt}
\begin{tabular}{rrrrrl}\toprule
group & shared & mean & MEB & zero & shared$-$best global\\\midrule
8   & 0.596 & 0.737 & 0.691 & 0.604 & -0.008 [-0.078, +0.041]\\
32  & 0.611 & 0.748 & 0.711 & 0.631 & -0.018 [-0.082, +0.026]\\
128 & 0.692 & 0.790 & 0.768 & 0.702 & -0.010 [-0.037, +0.009]\\
\bottomrule\end{tabular}\end{table}

\begin{table}[t]\centering\footnotesize
\caption{Paired comparison of shared minus each global-centre certificate (acquired fraction, witness
order; dataset-cluster bootstrap over the four datasets, pooled over encoders and groups).}\label{tab:acqcontrast}
\begin{tabular}{lrr}\toprule
baseline & mean difference & 95\% interval\\\midrule
mean & -0.125 & [-0.271, -0.022]\\
MEB  & -0.090 & [-0.190, -0.020]\\
zero & -0.013 & [-0.055, +0.021]\\
\bottomrule\end{tabular}\end{table}

The dataset-cluster intervals for shared minus mean-centred and shared minus MEB-centred lie below zero,
but shared is statistically indistinguishable from the zero-centred bound (\cref{tab:acqcontrast}). One
explanation, not tested here, is that in optional assignment many robust moves drop a target. For a drop
(owner $0$, $q_0=0$) the shared and zero-centred bounds coincide at $|\lambda|\,\|q_{a(j)}\|$, whereas a
centre $c\neq0$ gives $|\lambda|(\|c\|+\|q_{a(j)}-c\|)$.

\subsection{Runtime}
\begin{table}[t]\centering\footnotesize
\caption{Online wall time (seconds, median over cells of per-cell medians) for each selective variant
and its scope-matched eager baseline; ratio is the geometric mean over cells of selective/full (above 1
means slower). Acquired: fraction of active targets.}\label{tab:runtime}
\begin{tabular}{lrrrr}\toprule
variant & selective & full & ratio & acquired\\\midrule
shared/e2e/geo/g128   & 1.566 & 1.059 & 1.33 & 0.980\\
shared/known/geo/g128 & 1.255 & 0.733 & 1.53 & 0.980\\
zero/known/geo/g128   & 1.160 & 0.733 & 1.44 & 0.980\\
\bottomrule\end{tabular}\end{table}

\paragraph{The geometric schedule removes most selectivity.} The prespecified timing variants double the
group each unresolved round. They acquired 92\%--100\% of active targets in every dataset, including
Walmart--Amazon (92\%), where fixed-group certification needed only 7\%--11\%. The runtime comparison
therefore measures a schedule that encodes nearly everything plus certificate overhead. The wall time of
a fixed small-group selective run in the high-savings regime was not measured; that is the main open
runtime question. Eager end-to-end encoding of all active targets took Abt--Buy 0.65\,s, Amazon--Google
0.93\,s, Walmart--Amazon 5.43\,s, DBLP--ACM 1.20\,s (median).

For the prespecified primary contrast (shared, end-to-end, geometric from 128, versus full end-to-end),
the selective-minus-full wall time is +0.510\,s (per encoder and dataset, the difference of median
wall times, averaged within each dataset and then over the four datasets; dataset-cluster bootstrap 95\%
interval [+0.248, +0.772]), so the prespecified runtime claim is \textsc{contradicted} (\cref{sec:ledger}). Known-source
timing, reported separately, gives the same direction (\cref{tab:runtime}).

\subsection{Numerical portability across batch sizes}
\begin{table*}[t]\centering\small
\caption{Re-encoding all candidate-active targets at fixed microbatches (1, 8, 32, 64) relative to the
canonical cache: largest row L2 and coordinate drift, largest edge-score change, fraction of
microbatches with an identical optimal mapping, and fraction where each assignment remains optimal under
the other's scores. Every audited row had FP64 norm at most 1.}\label{tab:batch}
\begin{tabular}{lrrrrr}\toprule
cell & max L2 & max coord. & max $|\Delta s|$ & mapping eq. & bidir.\ optimal\\\midrule
Abt--Buy / BGE-small & 1.55e-03 & 9.60e-04 & 7.07e-05 & 1.00 & 1.00\\
Abt--Buy / MiniLM-L6 & 1.18e-03 & 2.86e-04 & 7.92e-05 & 0.00 & 0.00\\
Amazon--Google / BGE-small & 1.66e-03 & 9.83e-04 & 2.02e-04 & 0.00 & 0.00\\
Amazon--Google / MiniLM-L6 & 1.44e-03 & 3.79e-04 & 2.17e-04 & 0.25 & 0.25\\
DBLP--ACM / BGE-small & 1.68e-03 & 9.90e-04 & 1.84e-04 & 1.00 & 1.00\\
DBLP--ACM / MiniLM-L6 & 1.48e-03 & 3.82e-04 & 2.41e-04 & 1.00 & 1.00\\
Walmart--Amazon / BGE-small & 2.15e-03 & 1.21e-03 & 1.24e-04 & 1.00 & 1.00\\
Walmart--Amazon / MiniLM-L6 & 1.32e-03 & 3.74e-04 & 1.13e-04 & 1.00 & 1.00\\
\bottomrule\end{tabular}\end{table*}

Uniform re-encodings change features (\cref{tab:batch}), and the optimal assignment is bidirectionally
preserved at every tested microbatch in only 5 of 8 cells. In contrast, online acquisition mixes rows
from batches of different composition, and 45 of 120 online runs produced an incumbent that was exactly
optimal on its own realization yet sub-optimal by up to $2.49\times10^{-6}$ on the eager cache. The
assignment is therefore not batch-invariant, although the certificate is exact on every realized
instance.

\subsection{Prespecified claims}\label{sec:ledger}
Scored from these records, the prespecified claims receive the following labels:
assignment is invariant to tested batch sizes: \textsc{contradicted};
blocking preserves gold candidate recall: \textsc{supported};
canonical embeddings satisfy unit ball: \textsc{supported};
exact recovery with certificate: \textsc{supported};
exact recovery with certificate on same realization: \textsc{supported};
neural score improves over lexical only: \textsc{weakened};
selective acquisition improves end to end runtime: \textsc{contradicted};
shared certificate reduces acquisition vs global centers: \textsc{weakened};
task quality matches full model: \textsc{supported}.
With four dataset clusters, a \textsc{supported} label means every cluster agrees in sign, not that the
effect is precisely estimated.

\section{Discussion}\label{sec:discussion}
\paragraph{Runtime.} The runtime claim is \RuntimeStatus{}. Eager encoding of all active targets
takes a few seconds at most with these small encoders, and the prespecified geometric schedule acquired
almost every target, so the selective loop paid the certificate and solver overhead without saving
features. In the one regime with large savings (Walmart--Amazon, about 7\% of 17{,}646 active targets
needed at small fixed groups), selective wall time was not measured. That comparison, ideally with a more
expensive encoder, is the main open question and would need a new prespecified variant.

\paragraph{Acquisition.} Savings are strongly dataset-dependent: large for Walmart--Amazon, moderate for
Amazon--Google and negligible for Abt--Buy and DBLP--ACM, where nearly every target is needed. The shared
bound beats the mean and MEB centres but not the zero centre.

\paragraph{Task value of the costly feature.} After development-only tuning, the neural score changes
held-out F1 by about zero relative to an independently tuned lexical-only assignment. Exact selective
recovery reproduces the full model, but on these benchmarks the full model is not clearly better than the
cheap one.

\paragraph{Numerical portability.} The certificate is exact on the realized instance, but the optimal
assignment itself is not batch-invariant (three of eight cells under uniform re-encoding; 45 of 120 online
runs against an eager cache). Deployments that re-encode features must treat batch invariance as a
separate requirement; the certificate does not provide it.

\paragraph{Limitations.} The evidence covers four text benchmarks, two small frozen encoders and a
frozen blocker graph; no image data is used, so transfer to vision association problems is untested.
With four dataset clusters the bootstrap intervals are coarse.

\section*{LLM Usage Statement}
The authors developed the research ideas and proofs. AI tools provided coding, debugging, writing, and editing support, including polishing the exposition and presentation of the proofs. The authors reviewed and verified the resulting material and retain responsibility for the research questions, methods, proofs, experiments, interpretation, and final content.

{
    \small
    \bibliographystyle{ieeenat_fullname}
    \bibliography{references}
}

\appendix

\section{Proof of \texorpdfstring{\cref{prop:dominance}}{Proposition 1} (dominance at a fixed known set)}\label{sec:s-proof}
Fix the incumbent $M$ with owner map $a$ and the known set $K$. For a fixed centre $c$, the centre-$c$
certificate uses the robust gains $g^{c}_j(i)$ obtained from \cref{eq:gain} by replacing the unknown-target term
$|\lambda|\,\|q_i-q_{a(j)}\|_2$ with $|\lambda|\big(\|q_i-c\|_2+\|q_{a(j)}-c\|_2\big)$; known targets and
the incumbent owner ($g_j(a(j))=0$) are treated identically by both certificates.

\emph{Gains.} For $j\in K$ the two gains coincide. For $j\notin K$ and any candidate owner $i$
(including the dummy $i=0$ with $q_0=0$), the triangle inequality gives
$\|q_i-q_{a(j)}\|_2\le\|q_i-c\|_2+\|q_{a(j)}-c\|_2$, so
$g^{\mathrm{shared}}_j(i)\le g^{c}_j(i)$ entrywise.

\emph{Certificate values.} Both $\Gamma_{\mathrm{shared}}$ and $\Gamma_c$ are maxima over the same set of
owner maps $a'$ of optional assignments in $G$ (\cref{thm:cert}). For every such $a'$,
$\sum_j g^{\mathrm{shared}}_j(a'(j))\le\sum_j g^{c}_j(a'(j))\le\Gamma_c$; taking the maximum over $a'$
gives $\Gamma_{\mathrm{shared}}\le\Gamma_c$. Hence $\Gamma_c\le0$ implies $\Gamma_{\mathrm{shared}}\le0$:
whenever the centre-$c$ certificate certifies, the shared certificate does as well. \hfill$\square$

As stated in \cref{sec:cert}, this compares bounds at one fixed state; it says nothing about the number
of targets acquired along an adaptive trajectory, which is measured empirically.

\section{Per-dataset acquisition and eager encoding time}\label{sec:s-perdataset}
\Cref{tab:s-perdataset} restates, in table form, the per-dataset acquired fractions of the shared
certificate reported in \cref{sec:results} (simulations, witness order, mean over the two encoders), together
with the fraction acquired by the prespecified geometric timing schedule and the median eager end-to-end
encoding time of all active targets.

\begin{table}[t]\centering\footnotesize
\caption{Shared certificate: fraction of candidate-active targets acquired at fixed group sizes 8, 32
and 128, and median eager end-to-end encoding time of all active targets.}\label{tab:s-perdataset}
\begin{tabular}{lrrrr}\toprule
dataset & $g{=}8$ & $g{=}32$ & $g{=}128$ & eager (s)\\\midrule
Abt--Buy        & 0.86 & 0.89 & 1.00 & 0.65\\
Amazon--Google  & 0.51 & 0.51 & 0.67 & 0.93\\
Walmart--Amazon & 0.07 & 0.07 & 0.11 & 5.43\\
DBLP--ACM       & 0.95 & 0.97 & 0.99 & 1.20\\
\bottomrule\end{tabular}\end{table}

The prespecified geometric timing variants acquired 92\%--100\% of active targets in every dataset,
including Walmart--Amazon (92\%), where fixed-group certification needed only 7\%--11\%. Walmart--Amazon
has 17{,}646 active targets.

\section{Reproducibility record}\label{sec:s-repro}
The reported core run and the corrected pilot each executed one pinned code snapshot (snapshot A and
snapshot B, respectively). \Cref{tab:s-repro} lists the recorded SHA-256 prefixes of the protocol, the run
configurations and the run records, so that the material used for the reported numbers can be matched
unambiguously.

\begin{table}[t]\centering\footnotesize
\caption{Run record: code snapshots and SHA-256 prefixes.}\label{tab:s-repro}
\input{generated/repro.tex}
\end{table}

\end{document}

%% file: generated/macros.tex
\newcommand{\PilotCacheGap}{$2.49\times10^{-6}$}
\newcommand{\RuntimeStatus}{contradicted}

%% file: figures/fig1.tex
\begin{tikzpicture}[
  font=\footnotesize,
  src/.style={circle,draw=black,fill=black!12,minimum size=6.2mm,inner sep=0pt},
  tknown/.style={rectangle,draw=black,fill=black!55,text=white,minimum size=5.6mm,inner sep=0pt},
  tunk/.style={rectangle,draw=black,dashed,fill=white,minimum size=5.6mm,inner sep=0pt},
  cand/.style={draw=black!45,thin},
  inc/.style={draw=black,line width=1.3pt},
  box/.style={rectangle,draw=black,rounded corners=1.5pt,fill=black!4,align=center,
              inner sep=2.2pt,text width=#1},
  box/.default=5.2cm,
  arr/.style={-{Stealth[length=1.8mm]},thick},
  ptitle/.style={font=\footnotesize\bfseries,anchor=west}
]
\node[ptitle] at (0,5.55) {(a) Exact assignment, costly targets};
\node[font=\scriptsize,align=center] at (0.75,4.95) {sources $S$\\($q_i$ known)};
\node[font=\scriptsize,align=center] at (3.85,4.95) {targets $T$\\($z_j$ costly)};
\node[src] (s1) at (0.75,4.05) {$q_1$};
\node[src] (s2) at (0.75,3.05) {$q_2$};
\node[src] (s3) at (0.75,2.05) {$q_3$};
\node[tknown] (t1) at (3.85,4.25) {$z_1$};
\node[tunk]   (t2) at (3.85,3.45) {?};
\node[tknown] (t3) at (3.85,2.65) {$z_3$};
\node[tunk]   (t4) at (3.85,1.85) {?};
\draw[cand] (s1)--(t2); \draw[cand] (s2)--(t3); \draw[cand] (s3)--(t3);
\draw[cand] (s2)--(t4);
\draw[inc] (s1)--(t1); \draw[inc] (s2)--(t2); \draw[inc] (s3)--(t4);
\node[font=\scriptsize,anchor=west] at (4.2,3.45) {$\|z_j\|\le 1$};
\node[font=\scriptsize,anchor=west] at (4.2,1.85) {unknown};
\draw[inc] (0.1,1.25)--(0.55,1.25);
\node[font=\scriptsize,anchor=west] at (0.6,1.25) {incumbent $M$};
\draw[cand] (2.45,1.25)--(2.9,1.25);
\node[font=\scriptsize,anchor=west] at (2.95,1.25) {candidate edge};
\node[align=center] at (2.45,0.55)
  {$s_{ij}=\underbrace{b_{ij}}_{\text{cheap}}+\lambda\,q_i^{\top}\underbrace{z_j}_{\text{costly}}$};
\begin{scope}[xshift=5.85cm]
\node[ptitle] at (0,5.55) {(b) Certify, otherwise acquire};
\node[box] (b1) at (2.7,5.05) {incumbent $M$, unknown $z_j$ set to $0$};
\node[box] (b2) at (2.7,4.25) {robust gain of each reassignment, using\\
  $\lambda(q_i{-}q_{a(j)})^{\top}z_j\le|\lambda|\,\|q_i{-}q_{a(j)}\|$};
\node[box] (b3) at (2.7,3.3) {best alternative by a second sparse\\
  assignment: $\Gamma=\max_{M'}\sum_j g_j(a'(j))$};
\node[box=2.45cm,fill=white,line width=0.9pt] (yes) at (1.3,2.05)
  {$\Gamma\le0$: stop; $M$ is optimal for \emph{every} completion};
\node[box=2.45cm,fill=white] (no) at (4.1,2.05) {$\Gamma>0$: acquire the alternative's unknown targets};
\draw[arr] (b1)--(b2); \draw[arr] (b2)--(b3);
\draw[arr] (yes.north|-b3.south) -- (yes.north);
\draw[arr] (no.north|-b3.south) -- (no.north);
\draw[arr] (no.east) -- ++(0.22,0) |- (b1.east);
\coordinate (qi) at (0.4,0.3);
\coordinate (qa) at (2.1,0.2);
\coordinate (c)  at (1.4,0.95);
\draw[line width=1.1pt] (qi)--(qa);
\draw[dashed,thick] (qi)--(c)--(qa);
\fill (qi) circle (1.5pt) node[left] {$q_i$};
\fill (qa) circle (1.5pt) node[right] {$q_{a(j)}$};
\fill[white] (c) circle (1.7pt); \draw (c) circle (1.7pt) node[above] {$c$};
\node[font=\scriptsize,align=left,anchor=west] at (3.15,1.05)
  {solid: shared bound};
\node[font=\scriptsize,align=left,anchor=west] at (3.15,0.72)
  {dashed: fixed centre $c$};
\node[font=\scriptsize,align=left,anchor=west] at (3.15,0.39)
  {(zero, mean, MEB)};
\node[font=\scriptsize,anchor=west] at (3.15,0.06) {shared $\le$ any $c$ (Prop.~1)};
\end{scope}
\begin{scope}[xshift=12.1cm]
\node[ptitle] at (0,5.55) {(c) Audit on 4 text EM benchmarks};
\begin{axis}[
  at={(0.95cm,2.45cm)}, anchor=south west,
  width=4.85cm, height=3.15cm, scale only axis=false,
  ybar, bar width=4.6mm, ymin=0, ymax=1,
  ytick={0,0.5,1}, yticklabel style={font=\scriptsize},
  symbolic x coords={shared,mean,MEB,zero}, xtick={shared,mean,MEB,zero},
  xticklabel style={font=\scriptsize}, enlarge x limits=0.2,
  ylabel={acquired, $g{=}8$}, ylabel style={font=\scriptsize,yshift=-1mm},
  nodes near coords, nodes near coords style={font=\scriptsize},
  point meta=explicit symbolic, axis line style={black}, tick style={black},
  axis x line*=bottom, axis y line*=left,
  clip=false]
\addplot[draw=black,fill=black!65,bar shift=0pt] coordinates {(shared,0.596) [0.60]};
\addplot[draw=black,fill=black!15,bar shift=0pt] coordinates
  {(mean,0.737) [0.74] (MEB,0.691) [0.69] (zero,0.604) [0.60]};
\end{axis}
\node[font=\scriptsize,align=left,anchor=north west,text width=5.1cm] at (0.0,2.0)
  {$\bullet$ exact on own realization: 96/96 simulations,\\
   \phantom{$\bullet$} 120/120 online runs\\
   $\bullet$ shared, $g{=}8$: 0.07 (Walmart--Amazon) to 0.95\\
   \phantom{$\bullet$} (DBLP--ACM) of active targets\\
   $\bullet$ runtime: +0.510\,s slower (claim contradicted)\\
   $\bullet$ vs.\ separately batched cache: 45/120 runs\\
   \phantom{$\bullet$} differ (batch sensitivity, not violations)};
\end{scope}
\end{tikzpicture}

%% file: generated/repro.tex
\begin{tabular}{ll}\toprule item & value\\\midrule
core run & snapshot A\\
corrected pilot & snapshot B\\
protocol & \texttt{5d77a4435ad7d0c9}\\
core configuration & \texttt{9dd0740d100656a2}\\
core run record & \texttt{f27cc0ec0e7af72f}\\
pilot configuration & \texttt{38164fa20b8a56d6}\\
pilot run record & \texttt{29b81aec174cd264}\\
\bottomrule\end{tabular}